%% file: main.tex
\pdfoutput=1
\documentclass[11pt,a4paper]{article}

\usepackage[utf8]{inputenc}
\usepackage[T1]{fontenc}
\usepackage{amsmath,amssymb,amsfonts}
\usepackage{amsthm}
\usepackage{graphicx}
\usepackage{booktabs}
\usepackage{array}
\usepackage{multirow}
\usepackage{subcaption}
\usepackage{xcolor}
\usepackage{hyperref}
\usepackage{natbib}
\usepackage{algorithm}
\usepackage{algpseudocode}
\usepackage{url}

\newtheorem{theorem}{Theorem}[section]
\newtheorem{lemma}[theorem]{Lemma}
\newtheorem{proposition}[theorem]{Proposition}
\newtheorem{corollary}[theorem]{Corollary}

\newtheorem{assumption}[theorem]{Assumption}
\newtheorem{remark}[theorem]{Remark}

\newcommand{\acc}[1]{\mathrm{acc}@#1}
\newcommand{\mae}{\mathrm{MAE}}

\title{On the Fundamental Limits of Prefill-Based Decoding Length Prediction}
\author{%
  Haowen Shi\textsuperscript{1},
  Xuzhe Yan\textsuperscript{1},
  Fahao Chen\textsuperscript{2},
  Peng Li\textsuperscript{1}\\[4pt]
  \small\textsuperscript{1}Xi'an Jiaotong University, Xi'an, China\\
  \small\textsuperscript{2}Shandong University, Jinan, China\\[4pt]
  \small\texttt{HaowenShi@outlook.com}, \texttt{Zedekiah5571@outlook.com},\\
  \small\texttt{chenfh@ieee.org}, \texttt{pengli@xjtu.edu.cn}%
}
\date{}

\begin{document}

\maketitle

\begin{abstract}
Predicting how many tokens a large language model (LLM) will decode, using only prefill-phase hidden states, is a key primitive for inference scheduling. We show that current benchmarks conflate predictor quality with task hardness: when the target model, dataset, or sampling configuration changes, reported accuracy collapses even though the predictor is unchanged. We trace this instability to an overlooked error source---the inherent stochasticity of autoregressive decoding---and derive bounds that every prefill-side method must obey. Our analysis treats decoding as a stopped stochastic process. The per-step stopping probability of a response is not constant but ramps up over hundreds of steps, so the decode length inherits a central limit theorem from the many local decisions that determine where a response ends; this yields per-prompt normality with an explicit error term, together with an affine variance law $\sigma=\alpha\mu+b$ whose slope is the coefficient of variation of the number of response segments. Empirically, the normal law reproduces the per-prompt MAE and the central mass of the decode-length distribution to within a few percent, and the residual departures from normality (mild right skew, slight excess kurtosis) are the ones predicted by the sampling-noise correction. Under fixed model, data, and sampling conditions, a broad design space of predictors (MLP, EGTP, FiLM, soft-label, and Mixture-of-Experts variants) converges to similar performance close to the bound, confirming that \emph{system error dominates model error}. We argue for a benchmarking protocol that fixes the model, data, and sampling parameters and that reports relative-error metrics alongside absolute ones. Code and analysis tools are available at \url{https://github.com/ShwStone/DecodingPredict}.
\end{abstract}

\section{Introduction}\label{sec:intro}

Large language model (LLM) inference consists of two distinct phases: a compute-bound \emph{prefill} phase that processes the input prompt in parallel, and a memory-bound \emph{decode} phase that generates output tokens autoregressively \citep{kwon2023efficient, yu2022orca}. Because the decode phase dominates end-to-end latency and energy consumption \citep{zhong2024distserve}, knowing how many tokens will be decoded \emph{before} decoding begins is invaluable for batch scheduling, PD disaggregation, and resource provisioning \citep{shahout2024dont, qiu2024efficient, zheng2023response}.

Motivated by this need, a growing body of work proposes to predict decoding length from prefill-phase information. Early approaches train external proxy models on surface features \citep{qiu2024efficient}. More recent methods reuse the target LLM's own internal activations: \citet{shahout2024dont} schedule requests with intermediate-layer embeddings; \citet{piotrowski2025graph} build a graph convolutional regressor over layer-wise hidden states; and the concurrent EGTP framework \citep{xie2026predicting} introduces Entropy-Guided Token Pooling (EGTP) over on-the-fly hidden states, achieving state-of-the-art results on the newly released ForeLen benchmark.

\textbf{The fairness problem.} Despite these advances, we observe that published numbers are difficult to compare. A predictor may achieve MAE $\approx$ 80 on ForeLen's long-sequence split with Qwen2.5-7B, yet the same architecture degrades to MAE $>$ 150 on ShareGPT with Qwen3-32B, or to MAE $>$ 260 on RL-sampled data. The root cause is that the \emph{inherent randomness} of autoregressive generation---not predictor design---is the dominant error source. When a query has high decode-length variance, \emph{no} prefill-side predictor can forecast a single deterministic length with high accuracy, because the length itself is a random variable.

\textbf{Our contributions.} We make three contributions.
\begin{enumerate}
    \item \textbf{A theory of where the error comes from.} We model decoding as a stopping time with a non-constant hazard and show that the decode length is asymptotically normal, with an explicit Berry--Esseen error term (\S\ref{sec:theory}). The same analysis yields the affine variance law $\sigma=\alpha\mu+b$ and identifies the fitted slope as the coefficient of variation of the number of response segments. From the normal law we then derive bounds that hold for \emph{every} prefill-side predictor, including sharpened MAE and pass-rate bounds and an exact decomposition of the excess risk into a \emph{system} and a \emph{model} term.
    \item \textbf{Empirical validation, including where the law fails.} On 300 ShareGPT prompts with 200 decodes each, we test the normal approximation against the quantities that the bounds actually use: the per-prompt MAE matches the Gaussian value $\sigma\sqrt{2/\pi}$ to within $1.9\%$, and the central mass matches to within two percentage points (\S\ref{sec:validation}). The residual departures from normality (median skewness $+0.27$, excess kurtosis $+0.35$) are those predicted by the sampling-noise correction of our analysis. We also confirm that $\sigma$ follows the affine law in $\mu$, with coefficients $(\alpha, b)$ that vary markedly across models and post-training recipes (Table~\ref{tab:sigma_models}).
    \item \textbf{Architecture-agnostic evaluation.} We benchmark a wide range of predictor designs---MLP regression on concatenated hidden states, EGTP-style entropy pooling, FiLM conditioning, soft-label classification, and MoE routing---under \emph{fixed} model, data, and sampling settings. All methods converge to similar performance close to the theoretical bound (\S\ref{sec:experiments}).
\end{enumerate}

Finally, we argue that existing benchmarks such as ForeLen conflate predictor quality with task hardness, and we recommend a controlled evaluation protocol that reports relative-error metrics alongside absolute ones (\S\ref{sec:discussion}).

\section{Preliminaries and Related Work}\label{sec:related}

\paragraph{Problem formulation.} Let a prompt $x$ be tokenized into $x_{1:L}$, and let $\mathbf{h}^{(l)}_{1:L}$ denote the hidden states of layer $l$, for $l\in[L_{\mathrm{model}}]$, computed during prefill. Let $Y$ be the number of new tokens decoded before the model emits an end-of-sequence (EOS) token, and let $\hat{y} = f\bigl(x, \{\mathbf{h}^{(l)}_{1:L}\}_{l}\bigr)$ be an estimate of $Y$ that uses prefill-observable information only.

\paragraph{Metrics.} Following prior work \citep{xie2026predicting}, we report the mean absolute error ($\mae = \frac{1}{N}\sum_i |\hat{y}_i - Y_i|$) and the \emph{pass-rate} accuracy at tolerance $\tau$,
\begin{equation}
    \acc{\tau} \;:=\; \Pr\bigl(|\hat{y} - Y| < \tau\bigr),
\end{equation}
that is, the fraction of test prompts whose prediction error is below $\tau$ tokens. We report $\acc{100}$, $\acc{200}$, $\acc{500}$, and $x(p90)$, the smallest $\tau$ for which $\acc{\tau}\ge 90\%$.

\paragraph{Feature extraction.} Most embedding-based methods pool the layer-wise hidden states into a fixed-dimensional vector. Common strategies are mean pooling over layers and tokens, taking the last token of the last layer, or concatenating all layers \citep{shahout2024dont, xie2026predicting}. Consistent with prior observations on sentence embeddings \citep{jiang2023prompteol, lei2024meta}, we find that concatenating all layers gives the richest representation for this task.

\paragraph{EGTP and ForeLen.} \citet{xie2026predicting} propose Entropy-Guided Token Pooling (EGTP), which computes a per-token entropy $H_i$ from the next-token distribution and uses $\mathrm{softmax}(H)$ as attention weights to pool hidden states. They introduce ForeLen, a benchmark with three splits (long sequences, chain-of-thought reasoning, and RL sampling), and report MAE reductions of up to $29.16\%$ over baselines. EGTP is an appealing feature-engineering technique, but ForeLen evaluates different methods on \emph{different} model--dataset combinations, so the reported gains cannot be separated from differences in task difficulty.

\paragraph{Other predictors.} \citet{piotrowski2025graph} treat layer-wise hidden states as a graph and apply a GCN regressor. \citet{zheng2023response} predict the length with a lightweight side network inserted at inference time. \citet{qiu2024efficient} train a separate proxy model on the prompt text alone. Our work differs from all of these in targeting the \emph{fundamental limits} that they share.

\paragraph{Iterative and online prediction.} All of the above methods make a \emph{single} prediction before decoding begins. A complementary line of work updates the estimate as partial output becomes available. \citet{choi2025elis} propose ELIS, which feeds the prompt plus the already-generated tokens into a BGE-based predictor at fixed intervals (e.g.\ every 50 tokens) and shows that MAE decreases monotonically as more context is revealed. \citet{wu2023fastserve} take a different route: instead of predicting the length, they estimate the remaining time adaptively with a skip-join MLFQ scheduler that re-prioritizes jobs at every iteration. Neither approach contradicts our prefill-side bounds: they operate on a richer information set ($x$ \emph{and} $y_{<t}$) and therefore face a smaller effective $\sigma$. We discuss their implications in \S\ref{sec:iterative}.

\section{Theoretical Analysis of Prediction Error}\label{sec:theory}

This section isolates the part of the prediction error that no predictor can remove. We first describe decoding as a stopped stochastic process and record exact hazard identities (\S\ref{sec:hazard}). We then give a mechanistic model of how a response acquires its length, from which normality of the decode length follows as a theorem rather than as an assumption (\S\ref{sec:clt}). Finally we translate that law into bounds valid for \emph{every} prefill-side predictor (\S\ref{sec:bounds}) and identify what controls the per-prompt standard deviation (\S\ref{sec:sigma_decomp}). To keep the exposition short, all proofs and derivations are deferred to Appendix~\ref{app:proofs}: the main text states each result, gives a one-line proof idea where useful, and points to the full argument.

\subsection{Decoding Length as a Stopping Time}\label{sec:hazard}

Fix a prompt $x$, a target model, and a sampling configuration. Decoding draws $y_1,y_2,\dots$ from the model's next-token distribution and stops at
\begin{equation}
    Y \;=\; \min\{t\ge1:\; y_t \in \mathcal{S}\},
\end{equation}
where $\mathcal{S}$ is the set of stop tokens (EOS together with any configured stop strings). Hence $Y$ is a \emph{stopping time} of the sampling process, and its randomness is exactly the randomness of the sampled continuation. Its law is governed by the \emph{stopping probability} (discrete hazard) and the \emph{survival function}
\begin{equation}\label{eq:hazard}
    p_t \;=\; \Pr\bigl(Y=t\mid Y\ge t,\,x\bigr),
    \qquad
    S_t \;=\; \Pr\bigl(Y\ge t\mid x\bigr) \;=\; \prod_{s=1}^{t-1}\bigl(1-p_s\bigr),
\end{equation}
with $S_1=1$ and $\Pr(Y=t\mid x)=S_tp_t$. The hazard is itself random, because the sampled context decides how likely the model is to stop at step $t$.

Appendix~\ref{app:hazard} records the exact moment identities implied by \eqref{eq:hazard} and shows that a \emph{constant} hazard would make $Y$ geometric, with $\sigma/\mu\to1$ and hence an irreducible relative MAE of about $74\%$; the measured $\sigma/\mu\approx0.15$ (\S\ref{sec:validation}) rules that model out. Moreover, since the indicators $\mathbf{1}\{Y\ge t\}$ that sum to $Y$ are nested rather than independent, normality cannot be inherited from a central limit theorem over tokens: it has to come from a coarser level, which is the subject of \S\ref{sec:clt}.

\subsection{A Renewal Model and a Central Limit Theorem}\label{sec:clt}

A long response is not one uninterrupted decision: it is a concatenation of segments---sentences, list items, clauses, or reasoning steps---each completed before the model decides whether to continue.

\begin{assumption}[Segment (renewal) structure]\label{ass:renewal}
Conditionally on $x$, the response splits into $M$ segments of lengths $L_1,\dots,L_M$ such that
(i) given $M$ and $x$, the $L_i$ are i.i.d.\ with mean $m$, variance $v$, and finite third absolute central moment $\rho$;
(ii) $M$ is a stopping count of the segment-level continuation process, independent of the $L_i$, with $\mathbb{E}[M]=n$ and $\mathrm{Var}(M)=c^{2}n^{2}+c_1n$ for constants $c,c_1\ge0$.
Then $Y=\sum_{i=1}^{M}L_i$.
\end{assumption}

\begin{theorem}[Normality of the decode length]\label{thm:clt}
Under Assumption~\ref{ass:renewal}, let $\mu=mn$, $\sigma^{2}=vn+m^{2}\mathrm{Var}(M)$, and $\rho_M=\mathbb{E}|M-n|^{3}/\mathrm{Var}(M)^{3/2}$. Suppose moreover that $M$ satisfies a Berry--Esseen bound at rate $n^{-1/2}$, i.e.\ $\sup_u\bigl|\Pr\bigl((M-n)/\sqrt{\mathrm{Var}(M)}\le u\bigr)-\Phi(u)\bigr|\le C_M\rho_Mn^{-1/2}$. Then there is a constant $C$ such that
\begin{equation}\label{eq:clt}
    \sup_{u\in\mathbb{R}}\Bigl|\Pr\Bigl(\frac{Y-\mu}{\sigma}\le u\;\Big|\;x\Bigr)-\Phi(u)\Bigr|
    \;\le\;\frac{C}{\sqrt{n}}\Bigl(\rho_M+\frac{\sqrt{v}}{mc}\Bigr),
\end{equation}
and hence $(Y-\mu)/\sigma\Rightarrow\mathcal{N}(0,1)$ as $n\to\infty$.
\end{theorem}

\noindent Assumption~\ref{ass:renewal} is the only statistical assumption of the paper, and it is deliberately coarse: it constrains a segment count rather than token-level behaviour. Appendix~\ref{app:clt} proves the theorem, discusses the regimes $c=0$ (renewal count) and $c>0$ (proportionally fluctuating segment count), and derives the consequences used below: the affine variance law $\sigma=\alpha\mu+b$ with $\alpha=\mathrm{CV}(M)$ (Corollary~\ref{cor:affine}); the aggregation index $\sigma/\mu\ge c$, with $k:=(\mu/\sigma)^{2}$ effective stopping opportunities (Corollary~\ref{cor:cv}); and the Gaussian--geometric shape of the residual skew (Remark~\ref{rem:shape}). The stopping ramp of Fig.~\ref{fig:hazard} and the measurements of \S\ref{sec:validation} check the assumption and its consequences empirically.

\subsection{Lower Bounds on Prediction Error}\label{sec:bounds}

Theorem~\ref{thm:clt} justifies the working model
\begin{equation}\label{eq:normality}
    Y \mid x \;\sim\; \mathcal{N}\bigl(\mu(x), \sigma^2(x)\bigr),
\end{equation}
up to the correction quantified in \eqref{eq:clt}. We now derive what this law implies for \emph{any} predictor $\hat{y}=f(x)$ that sees prefill-phase information only, and is therefore independent of the decoding noise given $x$.

\begin{proposition}[Irreducible MAE]\label{prop:mae}
For $Y \sim \mathcal N(\mu,\sigma^2)$,
\begin{equation}\label{eq:mae_min}
    \inf_{\hat{y}\in\mathbb R}\ \mathbb E\bigl[|Y-\hat{y}|\bigr] \;=\; \mathbb E\bigl[|Y-\mu|\bigr] \;=\; \sigma\sqrt{\frac{2}{\pi}} \;\approx\; 0.7979\,\sigma ,
\end{equation}
and the infimum is attained uniquely at $\hat{y}=\mu$.
\end{proposition}

\begin{proposition}[Maximal pass-rate accuracy]\label{prop:acc}
For every tolerance $\tau>0$ and every $\hat{y}\in\mathbb R$,
\begin{equation}\label{eq:acc_max}
    \acc{\tau} \;=\; \Pr\bigl(|Y-\hat{y}| < \tau\bigr) \;\le\; 2\Phi\!\left(\frac{\tau}{\sigma}\right) - 1,
\end{equation}
where $\Phi$ is the standard normal CDF.
\end{proposition}

\noindent Proposition~\ref{prop:mae} follows by differentiating $\hat{y}\mapsto\mathbb E|Y-\hat{y}|$ and Proposition~\ref{prop:acc} from a layer-cake (rearrangement) argument over intervals of length $2\tau$; both extrema are attained uniquely at $\hat{y}=\mu$, so the statements concern the \emph{best possible} predictor rather than the mean predictor (Appendix~\ref{app:opt}).

Propositions~\ref{prop:mae} and~\ref{prop:acc} state that, once $\sigma(x)$ is fixed, \emph{no} prefill-side predictor can do better: the bounds hold equally for MLPs, graph regressors, entropy-weighted pooling, and mixtures of experts. Concretely, $\acc{100}\ge90\%$ for a single prompt requires $\sigma\le100/\Phi^{-1}(0.95)=61$ tokens and $\acc{200}\ge90\%$ requires $\sigma\le122$ tokens; above these thresholds high accuracy is \emph{impossible}, whatever the capacity or the amount of training data.

A test set aggregates prompts with different $\sigma$. Because $\sigma\mapsto2\Phi(\tau/\sigma)-1$ is strongly nonlinear, averaging the per-prompt bounds and inserting $\bar\sigma$ differ materially---$57.9\%$ versus $46.2\%$ for the test set of \S\ref{sec:results}; we therefore always report the average of the per-prompt bounds (the curvature of the map is computed in Appendix~\ref{app:opt}).

\begin{proposition}[Excess risk]\label{prop:excess}
For any predictor $\hat{y}=f(x)$,
\begin{align}
    \mathbb{E}\bigl[(Y-\hat{y})^2\bigr] &= \underbrace{\mathbb{E}\bigl[\sigma^2(x)\bigr]}_{\text{system error}} + \underbrace{\mathbb{E}\bigl[(\mu(x)-\hat{y})^2\bigr]}_{\text{model error}}, \label{eq:excess_mse}\\
    \mathbb{E}\bigl[|Y-\hat{y}|\bigr] &\le \mathbb{E}\bigl[\sigma(x)\sqrt{2/\pi}\bigr] + \mathbb{E}\bigl[|\mu(x)-\hat{y}|\bigr]. \label{eq:excess_mae}
\end{align}
\end{proposition}

Equation~\eqref{eq:excess_mse} is the law of total variance and \eqref{eq:excess_mae} the triangle inequality applied prompt-wise (Appendix~\ref{app:opt}). Thus the gap between a predictor and the bound is at most its distance to the oracle mean $\mu(x)$, and since training labels are single decodes rather than the means $\mu(x)$, the regression target carries noise of variance $\sigma^2(x)$, which slows learning at finite $N$---the residual gap observed in \S\ref{sec:results}.

\begin{corollary}[Relative-error metrics are the right currency]\label{cor:relative}
Under the affine law \eqref{eq:affine}, for long responses ($\mu\gg b/\alpha$),
\begin{equation}\label{eq:relative}
    \frac{\mae_{\min}}{\mu} \;\longrightarrow\; \alpha\sqrt{\frac{2}{\pi}} \;\approx\; 0.798\,\alpha,
    \qquad
    \acc{100}_{\max} \;\longrightarrow\; 2\Phi\!\left(\frac{100}{\alpha\mu}\right)-1 .
\end{equation}
\end{corollary}

Absolute MAE therefore grows without bound with the length scale of the data, whereas the relative error tends to a stable, architecture-independent constant fixed by the generation process---the formal justification for the reporting protocol of \S\ref{sec:fairness}.

\subsection{Estimating $\sigma$ from Finitely Many Decodes}\label{sec:sigma_est}

In practice $\sigma$ is unobserved: we see only $K$ decodes per prompt. With $Y_1,\dots,Y_K\sim\mathcal N(\mu,\sigma^2)$ i.i.d.\ the sample variance satisfies $(K-1)S^2/\sigma^2\sim\chi^2_{K-1}$, so requiring $S$ to estimate $\sigma$ within $10\%$ at $95\%$ confidence amounts to the two simultaneous inequalities
\begin{equation}\label{eq:chi2}
    \chi^2_{K-1,0.025} \ge 0.81(K-1)
    \quad\text{and}\quad
    \chi^2_{K-1,0.975} \le 1.21(K-1),
\end{equation}
which first hold at $K=193$ (Appendix~\ref{app:chi2}); we use $K=200$ decodes per prompt. All bounds reported below should therefore be read as accurate to about $\pm10\%$ in $\sigma$, hence $\pm10\%$ in the bound itself.

\subsection{What Controls $\sigma$?}\label{sec:sigma_decomp}

Empirically, $\sigma$ is strongly correlated with the mean length $\mu$. Across six models evaluated on ShareGPT prompts, the relation is well described by the affine model derived in Appendix~\ref{app:clt} (Corollary~\ref{cor:affine}):
\begin{equation}\label{eq:sigma_model}
    \sigma \;=\; \alpha\,\mu \;+\; b .
\end{equation}
Table~\ref{tab:sigma_models} reports $(\alpha,b)$ for each model. The spread is large: Qwen3-4B-Instruct is comparatively stable ($\alpha=0.1068$), whereas Qwen3-4B-no-thinking is not ($\alpha=0.2029$). DeepSeek-R1-distilled-thinking combines a low slope ($\alpha=0.1156$) with an unusually large intercept ($b=147.6$), reflecting the exploratory randomness intrinsic to chain-of-thought generation even when the final answer is short. Since $\alpha\approx\mathrm{CV}(M)$, the fitted slopes say that a long answer to a ShareGPT prompt is worth roughly $1/\alpha^2\approx25$--$90$ effectively independent stopping opportunities.

\begin{table}[t]
\centering
\caption{Per-model $\sigma$--$\mu$ law on ShareGPT (300 prompts $\times$ 200 decodes). Lower $\alpha$ and $b$ indicate more stable generation. By Corollary~\ref{cor:affine}, $\alpha=\mathrm{CV}(M)$ and $b=\kappa_1/(2\alpha)$.}
\label{tab:sigma_models}
\begin{tabular}{@{}lccc@{}}
\toprule
Model & $\alpha$ & $\mu$ & $b$ \\
\midrule
Qwen3-4B-no-thinking & 0.2029 & 814.25 & 5.56 \\
Qwen3-4B-Instruct & 0.1068 & 1443.28 & 31.25 \\
DeepSeek-R1-distilled-thinking & 0.1156 & 676.00 & 147.60 \\
Llama3-8B & 0.2042 & 555.23 & $-1.39$ \\
Qwen3-32B & 0.1486 & 907.92 & 27.36 \\
Qwen3-30B-A3B-Instruct & 0.1682 & 1479.78 & 12.18 \\
\bottomrule
\end{tabular}
\end{table}

\section{Empirical Validation}\label{sec:validation}

\subsection{Is the Decode Length Normal?}\label{sec:normality_emp}

We sample 300 prompts from ShareGPT and decode each $K=200$ times with Qwen3-32B, using the model's recommended sampling configuration (temperature $0.6$, top-$k$ $20$, top-$p$ $0.95$) so that the length law reflects deployment. Two of the 300 prompts are deterministic (all 200 decodes return the same 8-token answer); there $\sigma=0$ and every statement below holds trivially. The remaining 298 prompts yield 200 lengths each.

A single $p$-value is a poor summary at $K=200$, since tests reject minuscule deviations once the sample is large. We therefore report three complementary diagnostics: how often normality is rejected, the size and direction of the deviations (standardized skewness and excess kurtosis), and---most importantly---how accurately the Gaussian law reproduces the quantities that the bounds actually use.

\emph{Rejection rates.} Shapiro--Wilk does not reject normality at $\alpha=0.05$ for 79 of the 298 prompts ($26.5\%$), while Kolmogorov--Smirnov does not reject for 226 ($75.8\%$). The gap reflects how sensitive Shapiro--Wilk is to tiny deviations at $K=200$.

\emph{Effect sizes.} The deviations are small and systematic rather than arbitrary: the median standardized skewness is $+0.27$ ($70.8\%$ of prompts are right-skewed) and the median excess kurtosis is $+0.35$; $58\%$ of prompts have $|\text{skew}|<0.5$ and $69\%$ have $|\text{kurt}|<1$. Pooling all 59{,}600 standardized lengths $(Y-\hat\mu)/\hat\sigma$ gives skewness $0.34$, excess kurtosis $1.48$, and a Kolmogorov--Smirnov distance of $0.019$ from $\mathcal N(0,1)$. This is precisely the correction predicted by Remark~\ref{rem:shape}: a Gaussian bulk plus a small right-skewed sampling component. Solving \eqref{eq:skew} for the observed medians gives $r=0.60$ (from the skewness) and $r=0.57$ (from the kurtosis), i.e.\ a mutually consistent decomposition with about a quarter of the variance in sampling noise.

\emph{Accuracy where it matters.} The per-prompt mean absolute deviation in units of $\sigma$ is $\mathbb E|Y-\mu|/\sigma=0.783$ (median over prompts), against the Gaussian prediction $\sqrt{2/\pi}=0.798$: the bound of Proposition~\ref{prop:mae} is accurate to $1.9\%$ and slightly conservative. Likewise, the empirical probability mass within $\tau/\sigma\in\{0.625,1,1.28,1.645\}$ exceeds the Gaussian value by $1.1$, $1.7$, $2.0$ and $1.1$ percentage points, respectively, so the bound of Proposition~\ref{prop:acc} errs on the safe side at the tolerances of interest. The normal model is thus not a convenient fiction: it is accurate to a few percent exactly where the bounds rely on it. The most visible exception is a genuine outlier prompt (CV $0.61$, skewness $9.1$), a runaway generation that no prefill-side predictor can anticipate; more generally the distribution of $\sigma$ is right-tailed (10\% of prompts have $\sigma>300$), which is why we report both the mean $\sigma$ ($162.3$) and the median ($133.9$) whenever ceilings are quoted.

\begin{figure}[t]
    \centering
    \includegraphics[width=0.95\textwidth]{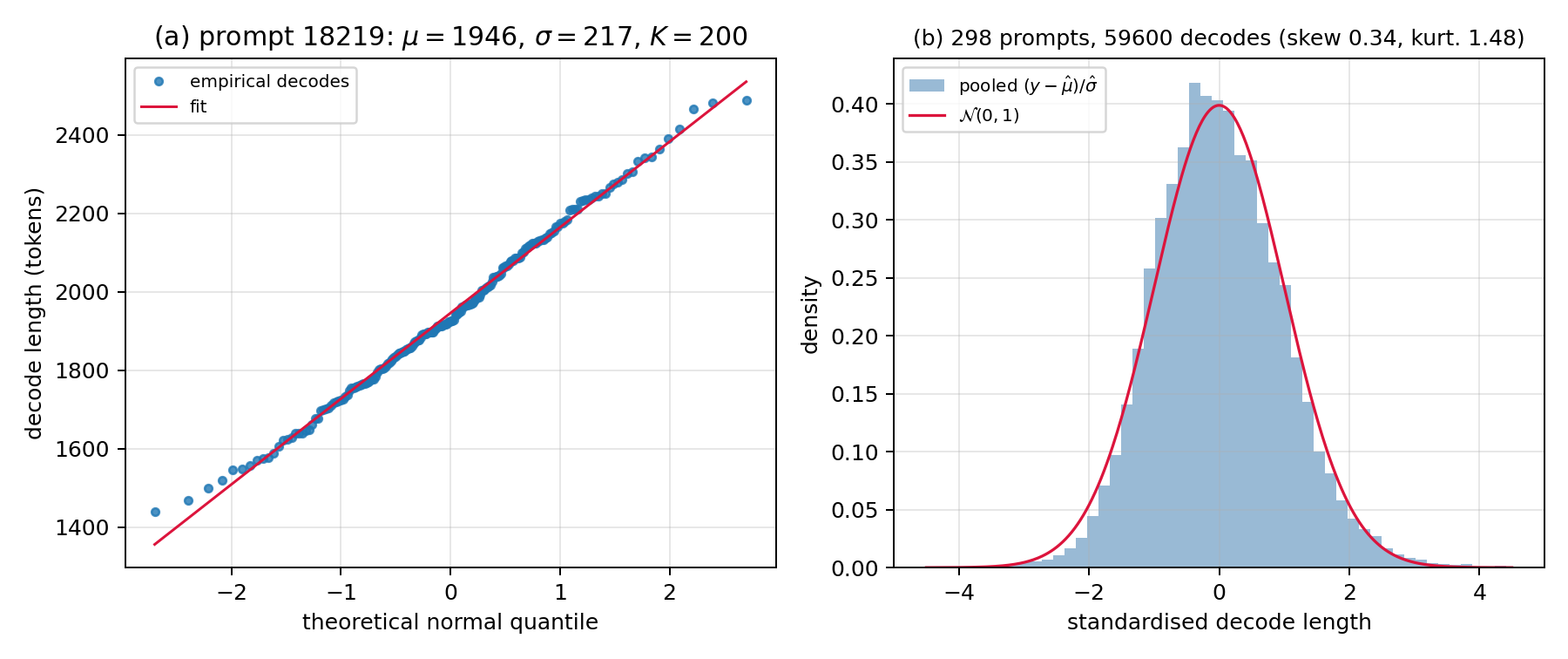}
    \caption{Normality of decode length (Qwen3-32B, ShareGPT). (a) Q--Q plot for a representative prompt ($\mu=1946$, $\sigma=217$, $K=200$); the deviations are confined to the extreme tails. (b) Pooled standardized lengths of all 298 non-degenerate prompts ($59{,}600$ decodes) against $\mathcal N(0,1)$: a slightly peaked, slightly right-skewed Gaussian, as predicted by Remark~\ref{rem:shape}.}
    \label{fig:qq}
\end{figure}

\begin{figure}[t]
    \centering
    \includegraphics[width=0.95\textwidth]{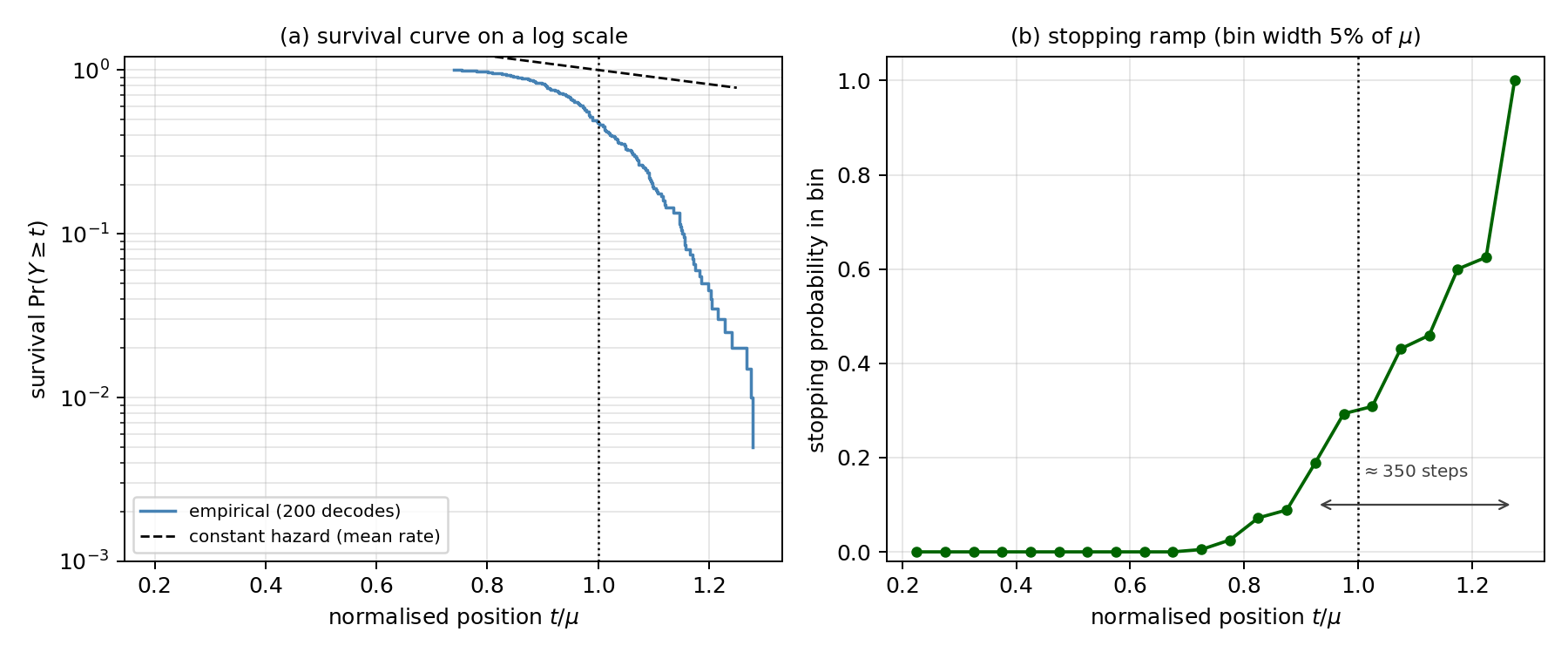}
    \caption{The stopping process of the prompt in Fig.~\ref{fig:qq}(a). (a) Empirical survival function on a log scale; a constant hazard would produce the dashed straight line, whereas the observed curve steepens with $t$, i.e.\ the stopping probability increases with length. (b) Bin-wise stopping probability (bin width $5\%$ of $\mu$): the hazard ramps from $\approx0$ to $\approx1$ over roughly 350 steps, which is the aggregation that Theorem~\ref{thm:clt} requires and Remark~\ref{rem:geometric} rules out.}
    \label{fig:hazard}
\end{figure}

Fig.~\ref{fig:qq}(a) shows the Q--Q plot of a representative long prompt and Fig.~\ref{fig:qq}(b) the pooled standardized law. Fig.~\ref{fig:hazard} displays the same prompt's survival curve on a log scale, which is markedly \emph{not} the straight line implied by a constant hazard, together with the stopping ramp that the CLT argument of \S\ref{sec:clt} relies on.

\subsection{$\mu$--$\sigma$ Correlation}

\begin{figure}[t]
    \centering
    \includegraphics[width=0.48\textwidth]{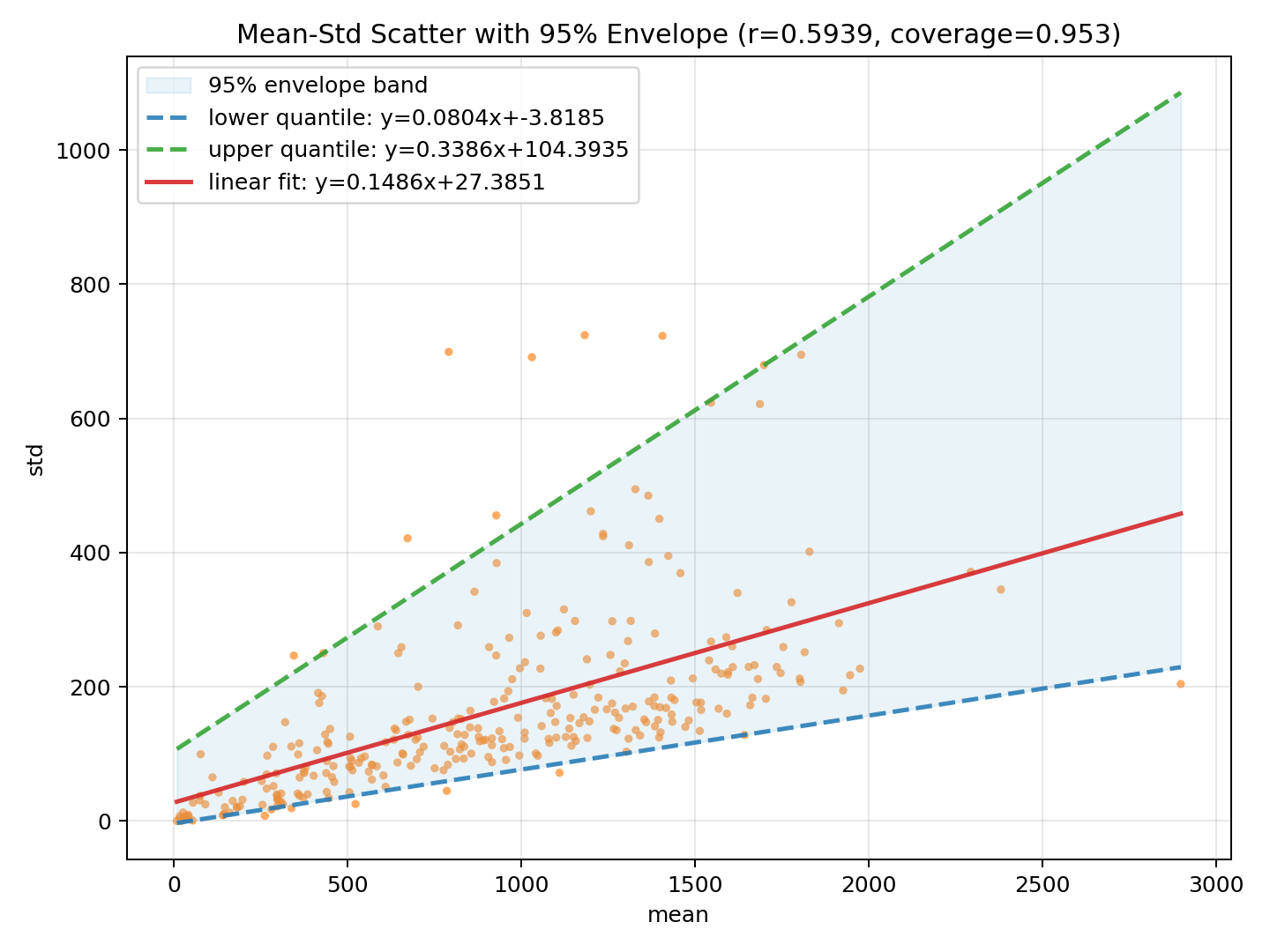}
    \caption{Per-prompt mean $\mu$ versus standard deviation $\sigma$ on ShareGPT (Qwen3-32B, 300 prompts $\times$ 200 decodes). The linear trend $\sigma=\alpha\mu+b$ of Corollary~\ref{cor:affine} is visible, with outliers reflecting prompt-specific constraints.}
    \label{fig:mean_std_corr}
\end{figure}

Figure~\ref{fig:mean_std_corr} plots per-prompt $(\mu,\sigma)$ pairs. The positive correlation is evident (Spearman $\rho=0.78$): longer expected outputs carry larger variance, as Corollary~\ref{cor:affine} predicts. Outliers above the trend line are open-ended questions (e.g.\ ``do you have reference text book\ldots''), while outliers below are tightly constrained instructions (e.g.\ code refactoring with prescribed variable names).

\subsection{MAE versus $\sigma$}

\begin{figure}[t]
    \centering
    \includegraphics[width=0.48\textwidth]{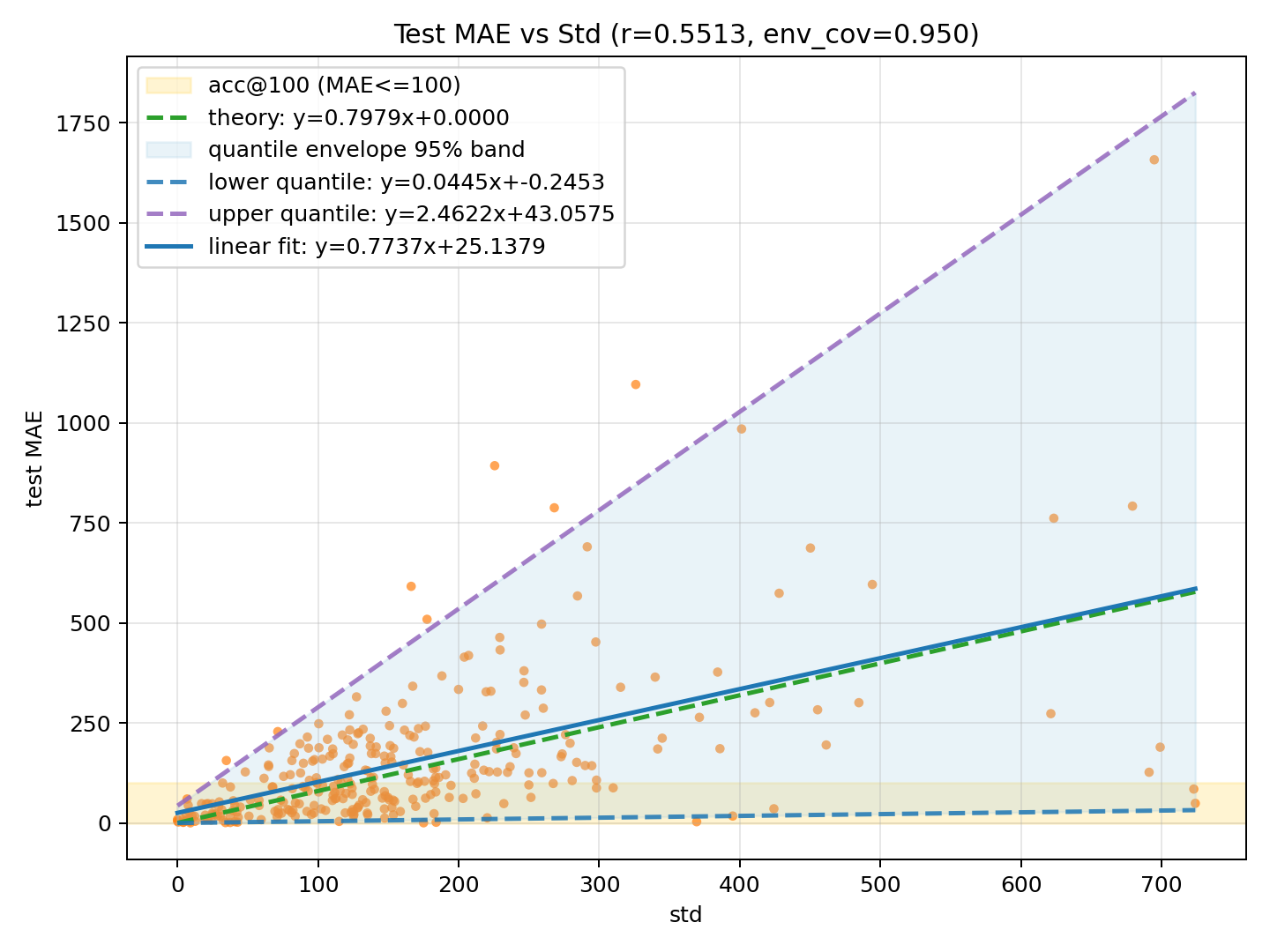}
    \caption{Per-prompt test MAE versus per-prompt $\sigma$ (Qwen3-32B, ShareGPT). The linear fit (blue) closely tracks the theoretical lower bound (red, $\mae_{\min}=\sigma\sqrt{2/\pi}$).}
    \label{fig:mae_vs_std}
\end{figure}

We train a simple 3-layer MLP on concatenated prefill hidden states and evaluate its per-prompt MAE on the same 300 ShareGPT prompts. Figure~\ref{fig:mae_vs_std} plots the test MAE against $\sigma$. The fitted slope is $0.774$, close to the theoretical slope $\sqrt{2/\pi}\approx0.798$ of Proposition~\ref{prop:mae}. In other words, the predictor is not the bottleneck: its error is the stochastic floor plus a small offset.

\section{Predictor Architectures and Results}\label{sec:experiments}

\subsection{Experimental Setup}

\textbf{Primary evaluation (ShareGPT self-test).} Our main experiments use ShareGPT as both training and test source. Prompts with IDs 0--18000 ($\approx$18K prompts) form the training and validation set, and prompts 18000--18300 (exactly 300) form the test set. All experiments fix the target model to Qwen3-32B (no-thinking mode) with its recommended sampling parameters (temperature $0.6$, top-$k$ $20$, top-$p$ $0.95$), so that generation matches deployment conditions. Prefill hidden states are exported once per prompt, and decode metadata is produced with the same model configuration. When evaluating other models (Llama3-8B, Qwen3-4B, DeepSeek-R1-distilled), we likewise adopt each model's own default sampling configuration.

\textbf{Why only 300 test prompts?} Testing our framework requires an accurate per-prompt estimate of $\sigma$. As shown in \S\ref{sec:sigma_est}, estimating $\sigma$ to within $10\%$ at $95\%$ confidence requires $K\ge193$ repeated decodes, and we use $K=200$; with 200 decodes per test prompt, the test set must stay small for the evaluation to be affordable. The training set, in contrast, uses single-shot decode labels, which suffice for fitting a predictor.

\textbf{Cross-dataset generalization.} Beyond the ShareGPT self-test we ran cross-dataset experiments (Appendix~\ref{app:cross}). For example, training on LMSYS-200K and testing on ShareGPT 18000--18300 yields MAE 182.40, and training on LeetCode and testing on HumanEval yields MAE 29.28 with $\acc{100}=96.28\%$. Predictor performance is thus highly sensitive to domain alignment between training and test, while the $\sigma$-driven ceiling remains the dominant constraint in every setting.

\textbf{Data generation caveat.} For resource reasons, prefill hidden states and length labels were produced by a PD-disaggregated pipeline: features were computed on prefill nodes and labels on separate decode workers. Batching, and hence low-level numerical behaviour, was not strictly controlled across the two stages; we return to this limitation in \S\ref{sec:discussion}.

\subsection{Architectures Evaluated}

We evaluate the following predictor designs, all of which see prefill hidden states only:

\begin{itemize}
    \item \textbf{MLP-Concat}: concatenate the last-token hidden vector of \emph{all} layers (${\sim}100$k dims) and regress through a 3-layer MLP (hidden width 128). This is our simplest baseline.
    \item \textbf{MLP-Mean}: mean-pool hidden states across layers and tokens, then MLP regression.
    \item \textbf{EGTP-style}: compute a per-token entropy from the next-token logits, softmax-normalize it into attention weights, and weighted-pool the hidden states \citep{xie2026predicting}; MLP regression on the pooled vector.
    \item \textbf{FiLM}: treat the prefill token count as a conditioning signal and use a FiLM generator \citep{perez2018film} to modulate the hidden representation before regression.
    \item \textbf{Soft-Label}: discretize lengths into 50 or 100 bins, train a classification head with Gaussian-smoothed targets, and map the expected bin back to a continuous length.
    \item \textbf{MoE-Hard}: train 4 expert regressors on length-based splits, with a hard gate classifier \citep{shazeer2017outrageously} selecting one expert.
    \item \textbf{MoE-Soft}: the same experts, with a soft gate that averages their predictions.
    \item \textbf{MoE-E2E}: end-to-end joint training of gate and experts with an auxiliary routing loss.
\end{itemize}

\subsection{Results}\label{sec:results}

Table~\ref{tab:results} reports MAE and pass-rate metrics. Several findings stand out.

\begin{table}[t]
\centering
\footnotesize
\setlength{\tabcolsep}{2pt}
\caption{Benchmark comparison under \emph{fixed} model (Qwen3-32B), dataset (ShareGPT 18000--18300), and sampling settings. All methods use prefill hidden states only. The Gaussian bound is the ceiling of Propositions~\ref{prop:mae} and~\ref{prop:acc}, averaged over the 300 test prompts, i.e.\ $\frac1N\sum_i\sigma_i\sqrt{2/\pi}$ and $\frac1N\sum_i[2\Phi(\tau/\sigma_i)-1]$; no prefill-side predictor can exceed it.}
\label{tab:results}
\resizebox{\linewidth}{!}{%
\begin{tabular}{@{}>{\raggedright\arraybackslash}p{2.2cm}ccccc@{}}
\toprule
Method & MAE $\downarrow$ & Acc@100 $\uparrow$ & Acc@200 $\uparrow$ & Acc@500 $\uparrow$ & x(p90) $\downarrow$ \\
\midrule
MLP-Mean & 181.21 & 45.07\% & 68.91\% & 93.51\% & 402.78 \\
MLP-Concat & 169.31 & 48.91\% & 71.11\% & 94.35\% & 391.09 \\
MLP-Concat + LayerNorm & 168.13 & 48.63\% & 71.65\% & 94.45\% & 392.13 \\
EGTP-style & 466.79 & 14.58\% & 25.35\% & 61.86\% & 891.48 \\
FiLM & 192.69 & 38.30\% & -- & -- & -- \\
Soft-Label (50 bins) & 183.24 & 45.05\% & -- & -- & -- \\
Soft-Label (100 bins) & 180.15 & 46.10\% & 69.83\% & 93.42\% & 413.24 \\
MoE-Hard & 177.91 & 46.54\% & 69.77\% & 93.35\% & 418.52 \\
MoE-Soft & 170.70 & 47.86\% & 71.00\% & 94.19\% & 393.70 \\
MoE-E2E & 171.39 & 47.33\% & 71.01\% & 94.30\% & -- \\
\midrule
\textbf{Theoretical bound} & \textbf{129.5} & \textbf{57.9\%} & \textbf{79.7\%} & \textbf{96.5\%} & -- \\
\bottomrule
\end{tabular}%
}
\end{table}

\textbf{Finding 1: simple is competitive.} MLP-Concat matches or beats FiLM, Soft-Label, and all MoE variants; the gap to the best MoE variant (MoE-Soft, MAE 170.70) is below $1\%$ relative. Notably, EGTP-style entropy-guided pooling \citep{xie2026predicting} degrades to MAE 466.79 on this split, far behind even the simplest MLP-Mean baseline, which suggests that entropy-weighted aggregation can inject noise when the token-level entropy distribution does not align with the length-predictive signal.

\textbf{Finding 2: the ceiling is tight and the residual gap is model error.} The mean per-prompt $\sigma$ is 162.3 (median 133.9; the $\sigma$ distribution is right-tailed, with $10\%$ of prompts above $\sigma=300$), so the Gaussian floor is $129.5$ tokens. Our best predictor achieves 168.1, an excess of 38.6 tokens. By Proposition~\ref{prop:excess} this excess is an upper bound on the model error $\mathbb E|\mu(x)-\hat y|$, i.e.\ at most $30\%$ of the system error $\mathbb E[\sigma(x)]\sqrt{2/\pi}$. Two effects account for it: training labels are single decodes and therefore carry noise of variance $\sigma^2(x)$ (Proposition~\ref{prop:excess}), and the test prompts are not perfectly aligned with the training distribution. Neither is architectural, which is consistent with architectures that differ by orders of magnitude in complexity landing within a few percent of each other.

\textbf{Finding 3: 90\% Acc@100 is unattainable, and the shortfall is quantifiable.} The per-prompt-averaged Gaussian ceiling is $57.9\%$ at $\tau=100$ ($79.7\%$ at $\tau=200$, $96.5\%$ at $\tau=500$). Measured Acc@100 tops out at $48.9\%$, i.e.\ 8.9 points (or $15\%$ relatively) below the ceiling, and measured Acc@500 reaches $94.5\%$ against a ceiling of $96.5\%$. Reporting accuracy \emph{relative to the ceiling} is therefore far more informative than the bare percentage (\S\ref{sec:fairness}). Inserting the mean $\bar\sigma=162.3$ into \eqref{eq:acc_max} instead of averaging per-prompt bounds would give $46.2\%$ and understate the ceiling by 12 points, which is why we always average the per-prompt bounds.

The same arithmetic explains the other models. Llama3-8B ($\bar\sigma=112$) and Qwen3-30B-A3B ($\bar\sigma=261$) face much easier and much harder tasks respectively; Table~\ref{tab:ceiling} tabulates the resulting ceiling once, as a function of $\sigma$, so that any predictor can be placed against it without re-deriving the bound.

\begin{table}[t]
\centering
\small
\setlength{\tabcolsep}{4pt}
\caption{Gaussian Acc@$\tau$ ceiling $2\Phi(\tau/\sigma)-1$ (Proposition~\ref{prop:acc}) as a function of the per-prompt standard deviation. A $90\%$ Acc@100 ceiling requires $\sigma\le61$ tokens.}
\label{tab:ceiling}
\begin{tabular}{@{}lccccccc@{}}
\toprule
$\sigma$ (tokens) & 60 & 100 & 150 & 200 & 250 & 300 & 400 \\
\midrule
Acc@100 ceiling & 90.4\% & 68.3\% & 49.5\% & 38.3\% & 31.1\% & 26.1\% & 19.7\% \\
Acc@200 ceiling & 99.9\% & 95.4\% & 81.8\% & 68.3\% & 57.6\% & 49.5\% & 38.3\% \\
Acc@500 ceiling & $>99.9$\% & $>99.9$\% & 99.9\% & 98.8\% & 95.4\% & 90.4\% & 78.9\% \\
\bottomrule
\end{tabular}
\end{table}

\section{Discussion}\label{sec:discussion}

\subsection{Toward Fairer Benchmarks}\label{sec:fairness}

The ForeLen benchmark \citep{xie2026predicting} evaluates different methods on different model--dataset combinations (e.g.\ Qwen2.5-7B on long sequences, Llama3.2-3B on RL sampling). Because each combination has its own $\sigma$, the reported MAE differences confound predictor quality with task hardness. We recommend that future benchmarks
\begin{enumerate}
    \item fix the target LLM, the dataset, and the sampling parameters when comparing predictors;
    \item report the per-prompt $\sigma$ (or at least its average and median over the test set), so that absolute MAE values can be read against the bound of Propositions~\ref{prop:mae} and~\ref{prop:acc};
    \item adopt \emph{relative-error} metrics (MAE/$\mu$, or pass rates expressed as a fraction of the ceiling $2\Phi(\tau/\sigma)-1$) alongside absolute MAE, since Corollary~\ref{cor:relative} shows that only the former is stable under changes of the length scale.
\end{enumerate}

\subsection{Runtime Compute Inconsistency}\label{sec:inconsistency}

Our prefill features and decode labels were produced by separate pipeline stages on different workers, without strict control of batch size, CUDA kernel versions, or deterministic mode. The hidden states used as features and those that produced the labels may therefore differ at the floating-point level. We believe this effect is small relative to the $\sigma$-driven system error, but it remains uncontrolled. Future work should generate features and labels in the \emph{same} forward pass, or at least fix all runtime determinism flags.

\subsection{PD-Disaggregated Generation}\label{sec:pd}

To obtain labels at scale we used a PD-disaggregated serving setup \citep{zhong2024distserve} in which prefill nodes export hidden states and KV caches while decode nodes generate completions. The separation introduces two potential biases: numerical differences between prefill and decode workers (as above), and possibly different batching strategies on the two sides. Since our training labels are single decodes rather than means of 200 decodes, label noise is amplified further. These are practical constraints of large-scale data collection; they should be acknowledged and, ideally, controlled in future studies.

\subsection{Iterative Prediction as Uncertainty Reduction}\label{sec:iterative}

Our analysis assumes that the predictor sees prefill-phase information only. This is the hardest regime, because $\mathrm{Var}(Y\mid x)$ is then as large as it can be. Once decoding begins, every generated token $y_t$ carries information about the remaining length, and the posterior variance can only shrink: $\mathrm{Var}(Y\mid x, y_{<t})\le\mathrm{Var}(Y\mid x)$, strictly so for most prompts. \citet{choi2025elis} exploit this by re-running a BGE-based predictor at fixed intervals during decoding and report that MAE decreases monotonically as more tokens are observed.

Theoretically, iterative prediction is the most effective way to reduce uncertainty without changing the generation process. Prefill-side feature engineering can only shrink the \emph{model error} of Proposition~\ref{prop:excess}, whereas iterative prediction shrinks the \emph{system error} by reducing the conditional variance of the target. In the limit where the predictor is invoked after every token, the remaining length collapses to a near-deterministic countdown and the MAE bound tends to zero.

The price is runtime cost: each prediction requires an encoder forward pass, adding latency and memory pressure to the serving system. For scheduling applications that consume prefill-side estimates to batch requests, this overhead is often acceptable only for coarse-grained updates (e.g.\ every 50--100 tokens). We therefore view prefill-only bounds and iterative refinement as complementary: the former set the baseline achievable without decode-side overhead, the latter quantify the uncertainty reduction available to systems that are willing to pay for it.

\section{Conclusion}\label{sec:conclusion}

We have shown that decoding-length prediction from prefill hidden states is fundamentally constrained by the stochasticity of autoregressive generation. Treating decoding as a stopping time and modelling the growth of a response as a renewal process yields per-prompt normality with an explicit error term, the affine variance law $\sigma=\alpha\mu+b$, and lower bounds on MAE and pass-rate accuracy that no prefill-side predictor can beat. Empirically these bounds are tight: a simple MLP already lands within $30\%$ of the noise floor, while more elaborate designs (EGTP, FiLM, Soft-Label, MoE) yield marginal or negative returns. The implication is that effort should shift from ever more complex predictors toward (i) \emph{controlling} the variance of the generation process itself---through lower-temperature sampling, constrained decoding, or prompt design---and (ii) benchmarking protocols that fix the model, data, and sampling parameters and report relative errors against the ceiling.

\paragraph{Code availability.} All code for data generation, model training, and evaluation, together with the prefill feature extraction utilities described above, is available at \url{https://github.com/ShwStone/DecodingPredict}.

\bibliographystyle{plainnat}
\bibliography{references}

\clearpage
\appendix
\input{appendix}

\end{document}

%% file: appendix.tex
\section{Detailed Derivations}\label{app:proofs}

\subsection{Hazard Identities and the Geometric Baseline}\label{app:hazard}

\begin{lemma}[Hazard identities]\label{lem:hazard}
For any hazard sequence $\{p_t\}_{t\ge1}$ with survival function $S_t=\prod_{s=1}^{t-1}(1-p_s)$ as in \eqref{eq:hazard},
\begin{equation}\label{eq:hazard_moments}
\begin{gathered}
    \mathbb E[Y]=\sum_{t\ge1}S_t,\qquad
    \mathbb E[Y^{2}]=\sum_{t\ge1}(2t-1)S_t,
    \\
    \mathrm{Var}(Y)=\sum_{t\ge1}(2t-1)S_t-\Bigl(\sum_{t\ge1}S_t\Bigr)^{2}.
\end{gathered}
\end{equation}
\end{lemma}

\paragraph{Proof.} For an integer-valued $Y\ge1$ the layer-cake identities read $Y=\sum_{t\ge1}\mathbf 1\{Y\ge t\}$ and, using $\sum_{t=1}^{n}(2t-1)=n^{2}$,
\begin{equation}
    Y^{2}=\sum_{t\ge1}(2t-1)\,\mathbf 1\{Y\ge t\}.
\end{equation}
Taking expectations and using $\mathbb E[\mathbf 1\{Y\ge t\}]=\Pr(Y\ge t)=S_t$ gives \eqref{eq:hazard_moments}. \hfill$\square$

\begin{remark}[A constant hazard gives an exponential law, not a normal one]\label{rem:geometric}
If $p_t\equiv p$ then $Y\sim\mathrm{Geom}(p)$ and the coefficient of variation is $\sqrt{1-p}\to1$, so the irreducible relative error is $\mathbb E|Y-\mu|/\mu\to2/e\approx0.74$ regardless of the output length. Since the indicators $\mathbf 1\{Y\ge t\}$ that sum to $Y$ are nested rather than independent, a first-passage time does not inherit the CLT of a sum: normality requires many nearly independent contributions, which enter at a coarser level than the token.
\end{remark}

\paragraph{Justification.} For $p_t\equiv p$ we have $S_t=(1-p)^{t-1}$, so \eqref{eq:hazard_moments} gives $\mathbb E[Y]=1/p$ and $\mathrm{Var}(Y)=(1-p)/p^{2}$. For small $p$ the rescaled variable $pY$ converges to a unit-rate exponential, whence
\begin{equation}
    \frac{\mathbb E|Y-\mathbb E Y|}{\mathbb E Y}\;\longrightarrow\;\mathbb E|E-1|=\frac{2}{e}\approx0.736,
    \qquad E\sim\mathrm{Exp}(1),
\end{equation}
where the constant follows from $\int_{0}^{1}(1-x)e^{-x}\,dx+\int_{1}^{\infty}(x-1)e^{-x}\,dx=2e^{-1}$. An $O(1)$ relative error is thus unavoidable under a constant hazard, no matter how much is known about $x$, which is why the measured $\sigma/\mu\approx0.15$ excludes that model. \hfill$\square$

\subsection{Proof of Theorem~\ref{thm:clt} and Its Consequences}\label{app:clt}

\paragraph{Wald identities.} Write $Y=\sum_{i=1}^{M}L_i$ with $M$ independent of the i.i.d.\ segments. Then $\mathbb E[Y]=m\,\mathbb E[M]=mn$, and by the law of total variance and Wald's identity $\mathbb E[\mathrm{Var}(Y\mid M)]=v\,\mathbb E[M]$, so
\begin{equation}\label{eq:wald}
    \mathrm{Var}(Y)=v\,n+m^{2}\,\mathrm{Var}(M)=v\,n+m^{2}\bigl(c^{2}n^{2}+c_1 n\bigr)
    \;=\;\kappa_2\mu^{2}+\kappa_1\mu,
\end{equation}
with $\kappa_2=c^{2}$ and $\kappa_1=v/m+m\,c_1$. Taking square roots and expanding,
\begin{equation}
    \sigma=\sqrt{\kappa_2\mu^{2}+\kappa_1\mu}
    =\alpha\mu\sqrt{1+\frac{\kappa_1}{\kappa_2\mu}}
    =\alpha\mu+\frac{\kappa_1}{2\alpha}+O(\mu^{-1}),
\end{equation}
This is the affine variance law:

\begin{corollary}[Affine variance law]\label{cor:affine}
Under Assumption~\ref{ass:renewal}, with $\kappa_2=c^{2}$ and $\kappa_1=v/m+m\,c_1$,
\begin{equation}\label{eq:affine}
    \sigma^{2}=\kappa_2\mu^{2}+\kappa_1\mu,
    \qquad\text{hence}\qquad
    \sigma=\alpha\mu+b+O\bigl(\mu^{-1}\bigr),
\end{equation}
where $\alpha=\sqrt{\kappa_2}=c$ and $b=\kappa_1/(2\alpha)$. The slope is therefore the coefficient of variation of the segment count and the intercept collects per-segment noise discounted by $1/(2\alpha)$---properties of the \emph{generation process} rather than of the predictor.
\end{corollary}

\begin{corollary}[Aggregation index]\label{cor:cv}
Under Assumption~\ref{ass:renewal},
\begin{equation}\label{eq:cv}
    \frac{\sigma}{\mu}=\sqrt{c^{2}+\frac{v/m^{2}+c_1}{n}}\;\ge\;c,
\end{equation}
with equality as $n\to\infty$ whenever $c>0$. Hence $k:=(\mu/\sigma)^{2}$ measures how many effectively independent stopping opportunities a prompt offers: models with a single deciding moment have $k=O(1)$, aggregated models have $k\gg1$. For the prompts of \S\ref{sec:validation} the median of $\sigma/\mu$ is $0.145$, i.e.\ $k\approx47$.
\end{corollary}

\paragraph{The two regimes.} The variance form $\mathrm{Var}(M)=c^{2}n^{2}+c_1n$ of Assumption~\ref{ass:renewal} covers both cases of interest. With $c=0$ the segment count behaves like a renewal count, $\mathrm{Var}(M)\propto n$, and the sampling and content terms of the decomposition below are of the same order. With $c>0$ the number of segments fluctuates proportionally across decodes of the same prompt---the regime implied by a fluctuating content path---and $c=\mathrm{CV}(M\mid x)$, which is the regime of the measurements in \S\ref{sec:validation}.
\paragraph{The two-scale decomposition.} With $A=\sum_{i\le M}(L_i-m)$ and $B=m(M-n)$ we have $Y-\mu=A+B$ exactly, with $\mathbb E[A]=\mathbb E[B]=0$, $\mathrm{Var}(A)=vn$ and $\mathrm{Var}(B)=m^2\mathrm{Var}(M)$, so $\sigma^2=\mathrm{Var}(A)+\mathrm{Var}(B)$. Although $A$ and $B$ are not independent (both depend on $M$), they are \emph{conditionally} independent given $M$, so by the tower property
\begin{equation}
    \varphi_{(Y-\mu)/\sigma}(t)
    =\mathbb E\Bigl[e^{itB/\sigma}\,\varphi_{A/\sigma}\bigl(t\mid M\bigr)\Bigr].
\end{equation}
Given $M=k$, the variable $A$ is a sum of $k$ centred i.i.d.\ terms, so a second-order expansion (or the Berry--Esseen bound for sums of length $k$) gives
\begin{equation}
    \Bigl|\varphi_{A/\sigma}\bigl(t\mid M=k\bigr)-1\Bigr|
    \le\frac{t^{2}}{2}\frac{\mathrm{Var}(A\mid M=k)}{\sigma^{2}}+O\!\Bigl(\frac{\rho\,|t|^{3}k^{3/2}}{m^{3}\sigma^{3}}\Bigr)
    =O\!\Bigl(\frac{t^{2}}{n}\Bigr),
\end{equation}
uniformly for $k=O(n)$, because $\mathrm{Var}(A\mid M=k)=vk$ and $\sigma^{2}\ge m^{2}c^{2}n^{2}$. Since $|e^{itB/\sigma}|=1$,
\begin{equation}
    \Bigl|\varphi_{(Y-\mu)/\sigma}(t)-\varphi_{B/\sigma}(t)\Bigr|
    \le\mathbb E\Bigl[\bigl|\varphi_{A/\sigma}(t\mid M)-1\bigr|\Bigr]=O\!\Bigl(\frac{t^{2}}{n}\Bigr),
\end{equation}
i.e.\ the sampling term perturbs the characteristic function only at order $1/n$. For the content term, the assumed Berry--Esseen bound for $M$ gives, with $\sigma_B=m\sqrt{\mathrm{Var}(M)}$,
\begin{equation}
    \Bigl|\varphi_{B/\sigma}(t)-e^{-t^{2}/2}\Bigr|
    \le \Bigl|\varphi_{B/\sigma_B}(t)-e^{-t^{2}/2}\Bigr|+\frac{t^{2}}{2}\Bigl|1-\frac{\sigma_B^{2}}{\sigma^{2}}\Bigr|
    \le C\Bigl(\rho_M\frac{|t|^{3}}{\sqrt n}+\frac{|t|^{2}}{n}\Bigr),
\end{equation}
where $1-\sigma_B^2/\sigma^2=vn/\sigma^2=O(n^{-1})$. Combining the three displays, for every fixed $t$,
\begin{equation}
    \Bigl|\varphi_{(Y-\mu)/\sigma}(t)-e^{-t^{2}/2}\Bigr|
    \le C'\Bigl(\rho_M\frac{|t|^{3}}{\sqrt n}+\frac{|t|^{2}}{n}\Bigr)\longrightarrow0,
\end{equation}
and Lévy's continuity theorem gives $(Y-\mu)/\sigma\Rightarrow\mathcal N(0,1)$. Esseen's smoothing lemma upgrades pointwise characteristic-function convergence to the uniform CDF bound \eqref{eq:clt}. \hfill$\square$

\paragraph{What the theorem does and does not assume.} The argument does not derive normality from scratch. It reduces normality of a \emph{token-level} stopping time to normality of a \emph{segment count} $M$---a much coarser object, with only tens of effective units---and it prices the reduction: the token-level (sampling) contribution enters the variance only at relative order $vn/\sigma^{2}=O(n^{-1})$, so its shape is irrelevant to leading order. In the renewal regime ($c=0$, $\mathrm{Var}(M)\propto n$) the two terms are of the same order and Anscombe's theorem for randomly stopped sums plays the role of the content CLT; the conclusion is unchanged whenever $M/n\to1$ in probability. The empirical checks of \S\ref{sec:validation} test the outcome rather than the assumption: the affine law \eqref{eq:wald}, the ratio $\mathbb E|Y-\mu|/\sigma$, and the skewness/kurtosis pair all match the model's predictions.
\subsection{Proofs and Auxiliary Computations for Section~\ref{sec:bounds}}\label{app:opt}

\paragraph{Proof of Proposition~\ref{prop:mae}.} Let $Z\sim\mathcal N(0,1)$ and $c=(\hat y-\mu)/\sigma$, so that $\mathbb E|Y-\hat y|=\sigma\psi(c)$ with $\psi(c)=\mathbb E|Z-c|$. Differentiating under the expectation,
\begin{equation}
    \psi'(c)=\mathbb E\bigl[-\operatorname{sgn}(Z-c)\bigr]
      =\Pr(Z<c)-\Pr(Z>c)=2\Phi(c)-1,
\end{equation}
which is negative for $c<0$, zero at $c=0$, and positive for $c>0$; hence $c=0$ is the unique minimizer. The value follows from $u=z^{2}/2$:
\begin{equation}
    \mathbb E|Z|=2\int_{0}^{\infty}z\,\frac{e^{-z^{2}/2}}{\sqrt{2\pi}}\,dz
    =2\int_{0}^{\infty}\frac{e^{-u}}{\sqrt{2\pi}}\,du=\sqrt{\frac{2}{\pi}}.
\end{equation}
For $Y\sim\mathcal N(\mu,\sigma^2)$ this gives $\mae_{\min}=\sigma\sqrt{2/\pi}\approx0.7979\sigma$. \hfill$\square$

\paragraph{Proof of Proposition~\ref{prop:acc}.} The density $f(y)=\sigma^{-1}\varphi((y-\mu)/\sigma)$ is symmetric about $\mu$ and non-increasing in $|y-\mu|$, so its superlevel sets $A_s=\{y:f(y)\ge s\}$ are intervals centred at $\mu$ with lengths $\ell(s)$. By the layer-cake representation and the fact that the intersection of two intervals of lengths $2\tau$ and $\ell(s)$ has length at most $\min\{2\tau,\ell(s)\}$,
\begin{equation}
    \Pr(Y\in I)=\int_{0}^{\infty}\bigl|I\cap A_s\bigr|\,ds
    \le\int_{0}^{\infty}\min\{2\tau,\ell(s)\}\,ds
    =\Pr\bigl(|\mu-Y|<\tau\bigr)
\end{equation}
for every interval $I$ of length $2\tau$; the equality case is $I=[\mu-\tau,\mu+\tau]$, i.e.\ $\hat y=\mu$. Taking $I=[\hat y-\tau,\hat y+\tau]$ gives \eqref{eq:acc_max}. \hfill$\square$

\paragraph{Proof of Proposition~\ref{prop:excess}.} For a predictor $\hat y=f(x)$ that is independent of the decoding noise given $x$, the tower property gives
\begin{equation}
\begin{aligned}
    \mathbb E\bigl[(Y-\hat y)^{2}\bigr]
    &=\mathbb E\Bigl[\mathbb E\bigl[(Y-\mu(x))^{2}\mid x\bigr]\Bigr]
      +\mathbb E\bigl[(\mu(x)-\hat y)^{2}\bigr] \\
    &=\mathbb E\bigl[\sigma^{2}(x)\bigr]+\mathbb E\bigl[(\mu(x)-\hat y)^{2}\bigr],
\end{aligned}
\end{equation}
which is \eqref{eq:excess_mse} and identifies the two terms as the irreducible (system) and the reducible (model) error. For the absolute loss, the triangle inequality together with Proposition~\ref{prop:mae} applied prompt-wise, $\mathbb E[|Y-\hat y|\mid x]\ge\sigma(x)\sqrt{2/\pi}$, gives
\begin{equation}
    \mathbb E\bigl[|Y-\hat y|\bigr]
    \le\mathbb E\bigl[\sigma(x)\sqrt{2/\pi}\bigr]+\mathbb E\bigl[|\mu(x)-\hat y|\bigr],
\end{equation}
i.e.\ \eqref{eq:excess_mae}: the excess over the bound is at most the $L_1$ distance from the predictor to the oracle mean. \hfill$\square$

\paragraph{Proof of Corollary~\ref{cor:relative}.} Under \eqref{eq:affine}, $\sigma/\mu=\alpha\sqrt{1+\kappa_1/(\kappa_2\mu)}\to\alpha$, so Proposition~\ref{prop:mae} gives $\mae_{\min}/\mu\to\alpha\sqrt{2/\pi}$ and Proposition~\ref{prop:acc} gives $\acc{\tau}_{\max}=2\Phi(\tau/\sigma)-1\to2\Phi(\tau/(\alpha\mu))-1$. The bound therefore depends on $\mu$ only through the ratio $\tau/\mu$, which is what makes relative-error metrics scale-free. \hfill$\square$

\paragraph{Test-set aggregation.} For $g(\sigma)=2\Phi(\tau/\sigma)-1$,
\begin{equation}
    g'(\sigma)=-\frac{2\tau\,\varphi(\tau/\sigma)}{\sigma^{2}},
    \qquad
    g''(\sigma)=\frac{2\tau\,\varphi(\tau/\sigma)}{\sigma^{3}}\Bigl(2-\frac{\tau^{2}}{\sigma^{2}}\Bigr),
\end{equation}
so $g$ is decreasing everywhere, convex for $\sigma>\tau/\sqrt{2}$ and concave below. Prompts with $\sigma\ll\tau$ saturate $g$ near $1$, so a test-set average is dominated by the convex region and typically exceeds $g(\bar\sigma)$: for the test set of \S\ref{sec:results} the per-prompt average is $57.9\%$ against $46.2\%$ for $g(\bar\sigma)$. This is why all ceilings in this paper average the per-prompt bounds.

\subsection{The Shape of the Deviations}\label{app:shape}

\begin{remark}[Variance budget and the shape of the deviations]\label{rem:shape}
Write the two terms of the decomposition
\begin{equation}\label{eq:twoscale}
    Y-\mu=\underbrace{\sum_{i=1}^{M}\bigl(L_i-m\bigr)}_{\text{sampling noise}}+\underbrace{m\bigl(M-n\bigr)}_{\text{content noise}}
\end{equation}
as $A$ and $B$, and let $r=\sigma_{\mathrm{samp}}/\sigma_{\mathrm{cont}}$. If the path-wise hazard is locally flat around the crossing point, $A$ behaves like a centred geometric variable, so the standardized law is a Gaussian--geometric convolution whose cumulants give
\begin{equation}\label{eq:skew}
    \mathrm{skew}(Y)=\frac{2r^{3}}{(1+r^{2})^{3/2}},
    \qquad
    \mathrm{kurt}(Y)=\frac{6r^{4}}{(1+r^{2})^{2}},
\end{equation}
with $\mathrm{kurt}$ the excess kurtosis; $r\to0$ recovers the Gaussian. The empirical medians (skewness $0.27$, excess kurtosis $0.35$; \S\ref{sec:validation}) invert to $r=0.60$ and $r=0.57$---mutually consistent, and implying that at the segment counts realized here ($n\approx45$, far from the asymptotic regime of Theorem~\ref{thm:clt}) about a quarter of the per-prompt variance is sampling noise. In this precise sense the decode length ``is'' Gaussian: the location and scale of the bound are Gaussian, and the residual right skew is a small, quantifiable correction rather than a modelling failure.
\end{remark}

\paragraph{Derivation.} Let $X=N+r\,(E-1)$ with $N\sim\mathcal N(0,1)$, $E\sim\mathrm{Exp}(1)$ independent, and $r\ge0$. Cumulants are additive, and the exponential law has $\kappa_3=2$, $\kappa_4=6$, so
\begin{equation}
    \mathrm{Var}(X)=1+r^{2},\qquad
    \kappa_3(X)=2r^{3},\qquad
    \kappa_4(X)=6r^{4},
\end{equation}
whence
\begin{equation}
    \mathrm{skew}(X)=\frac{2r^{3}}{(1+r^{2})^{3/2}},
    \qquad
    \mathrm{kurt}(X)=\frac{6r^{4}}{(1+r^{2})^{2}},
\end{equation}
which is \eqref{eq:skew}. The map $r\mapsto\mathrm{skew}(X)$ is strictly increasing, so $r$ can be inverted from an observed skewness (or, redundantly, from an observed excess kurtosis); the ratio $r^{2}/(1+r^{2})$ is the share of the variance carried by the geometric component. For the empirical medians of \S\ref{sec:validation} ($\text{skew}=0.27$, $\text{kurt}=0.35$) the two inversions give $r=0.60$ and $r=0.57$, i.e.\ a sampling-noise share of 24--26\%. Theorem~\ref{thm:clt} makes the sampling component vanish only in the limit $n\to\infty$; at the segment counts realized here ($n\approx45$) its share is small but not negligible, which is precisely why the residual skew is measurable.

\subsection{$\chi^{2}$ Confidence Bound for $\sigma$}\label{app:chi2}

Given $K$ i.i.d.\ observations $Y_1,\dots,Y_K\sim\mathcal N(\mu,\sigma^{2})$ with sample variance $S^{2}=\frac{1}{K-1}\sum_{i}(Y_i-\bar Y)^{2}$,
\begin{equation}
    \frac{(K-1)S^{2}}{\sigma^{2}}\sim\chi^{2}_{K-1}.
\end{equation}
We want $S$ to estimate $\sigma$ within $10\%$ relative error at confidence $0.95$, i.e. $\Pr(0.9\sigma\le S\le1.1\sigma)\ge0.95$, which is equivalent to
\begin{equation}
    \Pr\bigl(0.81(K-1)\le\chi^{2}_{K-1}\le1.21(K-1)\bigr)\ge0.95 .
\end{equation}
Let $\chi^{2}_{K-1,\alpha}$ denote the $\alpha$-quantile of the chi-squared distribution with $K-1$ degrees of freedom. Using the two central quantiles $0.025$ and $0.975$,
\begin{equation}
    \chi^{2}_{K-1,0.025}\ge0.81(K-1)
    \quad\text{and}\quad
    \chi^{2}_{K-1,0.975}\le1.21(K-1).
\end{equation}
Both inequalities hold simultaneously for the first time at $K=193$: there $\chi^2_{192,0.025}=155.52\ge0.81\cdot192=155.52$ and $\chi^2_{192,0.975}=232.27\le1.21\cdot192=232.32$, whereas at $K=192$ the corresponding values are $154.62<154.71$ and $231.16>231.11$. We round up to $K=200$ decodes per prompt. This requirement is the reason our test set is limited to 300 prompts: 200 decodes per prompt make an exhaustive evaluation of a larger test set prohibitively expensive.

\section{Training Details}\label{app:training}

All predictors are implemented in PyTorch and trained with Adam. Unless noted otherwise:
\begin{itemize}
    \item learning rate $10^{-3}$ with cosine decay;
    \item batch size 32;
    \item up to 100 epochs, with early stopping on validation MAE (patience 10);
    \item inputs are prefill hidden states exported with\linebreak \texttt{attn\_implementation="eager"};
    \item loss: Smooth-L1 (Huber) for regression, and a cross-entropy plus MSE mixture for soft labels.
\end{itemize}

\noindent\textbf{MLP-Concat.} Concatenate the last-token hidden vectors of all layers. For Qwen3-32B (64 layers, hidden size 5120) the input dimension is $327{,}680$, and the backbone applies $327680\to128\to128\to1$ linear maps with ReLU activations in between.

\noindent\textbf{EGTP-style.} Compute the entropy $H_i$ of the next-token distribution at each prefill position, normalise to weights $w_i=\mathrm{softmax}(H)_i$, weight-pool the hidden states per layer, and mean-pool across layers. An MLP regressor is fitted on the resulting 5120-dimensional vector.

\noindent\textbf{FiLM.} Condition on the prefill token count $L$ through a small MLP that produces a scale $\gamma$ and a shift $\beta$, applied element-wise to the mean-pooled hidden state before the regression head.

\noindent\textbf{Soft-Label.} Discretize target lengths into 100 equal-width bins and smooth the target with a Gaussian kernel of $\sigma_{\mathrm{bin}}=2$ bins. The loss is $\mathcal L=\lambda\mathcal L_{\mathrm{CE}}+(1-\lambda)\mathcal L_{\mathrm{MSE}}$ with $\lambda=0.5$.

\noindent\textbf{MoE variants.} Four experts cover four length intervals with $20\%$ overlap, namely $[0,375)$, $[375,750)$, $[750,1149)$ and $[1149,\infty)$. Each expert is a 2-layer MLP, and the gate is a 2-layer network with softmax output. MoE-E2E adds an auxiliary routing loss derived from the expert prediction intervals.

\section{Full Benchmark Results}\label{app:full_results}

Table~\ref{tab:full_results} reports mean and standard deviation over 10 independent training runs.

\begin{table}[h]
\centering
\caption{Mean $\pm$ std over 10 runs (Qwen3-32B, ShareGPT).}
\label{tab:full_results}
\begin{tabular}{@{}lcccc@{}}
\toprule
Method & MAE & Acc@100 & Acc@200 & Acc@500 \\
\midrule
MLP-Concat & $169.3 \pm 4.2$ & $48.9 \pm 1.1\%$ & $71.1 \pm 0.9\%$ & $94.4 \pm 0.3\%$ \\
EGTP-style & $466.8 \pm 1.1$ & $14.6 \pm 1.0\%$ & $25.4 \pm 3.0\%$ & $61.9 \pm 3.3\%$ \\
MoE-Soft & $170.7 \pm 6.3$ & $47.9 \pm 1.5\%$ & $71.0 \pm 1.2\%$ & $94.2 \pm 0.5\%$ \\
\bottomrule
\end{tabular}
\end{table}

\section{Cross-Dataset Generalization}\label{app:cross}

Table~\ref{tab:cross_dataset} summarizes experiments that train on one dataset and test on another (all with Qwen3-32B unless noted otherwise). Absolute MAE is strongly influenced by the alignment between the training and test domains, while the $\sigma$-driven ceiling stays the dominant constraint.

\begin{table}[h]
\centering
\footnotesize
\setlength{\tabcolsep}{2pt}
\caption{Cross-dataset generalization (train $\to$ test).}
\label{tab:cross_dataset}
\resizebox{\linewidth}{!}{%
\begin{tabular}{@{}>{\raggedright\arraybackslash}p{3.0cm}ccccc@{}}
\toprule
Train $\to$ Test & MAE & Acc@100 & Acc@200 & Acc@500 & x(p90) \\
\midrule
LMSYS-200K $\to$ ShareGPT-100K & 172.60 & 47.22\% & 70.89\% & 94.13\% & 393.05 \\
LMSYS-200K $\to$ ShareGPT-18k-18.3k & 182.40 & 43.75\% & 68.44\% & 92.43\% & 435.48 \\
LMSYS-200K $\to$ GSM8K & 44.33 & 92.96\% & 99.04\% & 99.88\% & 86.23 \\
ShareGPT-100K $\to$ ShareGPT-10K & 132.70 & 57.76\% & 80.16\% & 96.68\% & 299.96 \\
LeetCode $\to$ HumanEval & 29.28 & 96.28\% & 100.00\% & 100.00\% & 76.17 \\
\bottomrule
\end{tabular}%
}
\end{table}

When the domains are closely aligned (LeetCode $\to$ HumanEval, both code generation), the predictor achieves low MAE and high accuracy. When they differ substantially (LMSYS $\to$ ShareGPT), MAE drifts toward the ceiling implied by the test-set $\sigma$. The pattern reinforces the central claim: once the predictor is adequately trained, the residual error is dominated by the inherent stochasticity of generation rather than by architectural limitations.

\section{MoE Meta-Gate Definitions}\label{app:moe}

For completeness we record the confidence-aware meta gate used in the MoE experiments. Let $K=4$ experts, where expert $i$ outputs a mean prediction $\mu_i(x)$ and a risk estimate $r_i(x)$:
\begin{align}
    \mu_i(x) &= W_{i,\mu}\,h_i(x) + b_{i,\mu}, \\
    r_i(x) &= \mathrm{Softplus}\bigl(W_{i,r2}\,\mathrm{ReLU}(W_{i,r1}\,h_i(x) + b_{i,r1}) + b_{i,r2}\bigr).
\end{align}
The risk head is trained to predict the expert's absolute error $e_i(x,Y)=|\mu_i(x)-Y|$ with a Smooth-L1 loss.

The meta gate concatenates the feature $x$, the expert predictions $\mu(x)$, the risks $r(x)$, and the interval distances $d(x)$, where $d_i(x)=\max(a_i-\mu_i,0)+\max(\mu_i-b_i,0)$ for the expert interval $[a_i,b_i]$. It outputs a prior $\pi(x)=\mathrm{softmax}(g(z(x)))$ and scores
\begin{equation}
    s_i(x)=\log\bigl(\pi_i(x)+\varepsilon\bigr)-\alpha\,r_i(x)-\beta\,d_i(x),
\end{equation}
with routing weights $w_i(x)=\exp(s_i(x)/T)\big/\sum_j\exp(s_j(x)/T)$ and final prediction $\hat y(x)=\sum_i w_i(x)\,\mu_i(x)$.

The total training loss is $\mathcal L=\mathcal L_{\mathrm{reg}}+\lambda_{\mathrm{kl}}\mathcal L_{\mathrm{kl}}+\lambda_{\mathrm{bal}}\mathcal L_{\mathrm{bal}}$, where $\mathcal L_{\mathrm{reg}}$ is Smooth-L1 on $\hat y$, $\mathcal L_{\mathrm{kl}}$ is the KL divergence from the oracle soft routing targets $q_i\propto\exp(-|\mu_i-Y|)$, and $\mathcal L_{\mathrm{bal}}$ penalises unbalanced expert usage.

%% file: references.bib
@article{xie2026predicting,
  title={Predicting {LLM} Output Length via Entropy-Guided Representations},
  author={Xie, Huanyi and Chen, Yubin and Wang, Liangyu and Hu, Lijie and Wang, Di},
  journal={arXiv preprint arXiv:2602.11812},
  year={2026}
}

@article{shahout2024dont,
  title={Don't Stop Me Now: Embedding Based Scheduling for {LLMs}},
  author={Shahout, Rana and Malach, Eran and Liu, Chunwei and Jiang, Weifan and Yu, Minlan and Mitzenmacher, Michael},
  journal={arXiv preprint arXiv:2410.01035},
  year={2024}
}

@article{qiu2024efficient,
  title={Efficient Interactive {LLM} Serving with Proxy Model-Based Sequence Length Prediction},
  author={Qiu, Haoran and Mao, Weichao and Patke, Archit and Cui, Shengkun and Jha, Saurabh and Wang, Chen and Franke, Hubertus and Kalbarczyk, Zbigniew T and Basar, Tamer and Iyer, Ravishankar K},
  journal={arXiv preprint arXiv:2404.08509},
  year={2024}
}

@article{zheng2023response,
  title={Response Length Perception and Sequence Scheduling: An {LLM}-Empowered {LLM} Inference Pipeline},
  author={Zheng, Zangwei and Ren, Xiaozhe and Xue, Fuzhao and Luo, Yang and Jiang, Xin and You, Yang},
  journal={Advances in Neural Information Processing Systems},
  volume={36},
  pages={65517--65530},
  year={2023}
}

@inproceedings{piotrowski2025graph,
  title={Graph-Based Approach to Predicting Remaining Output Length in Large Language Models},
  author={Piotrowski, Grzegorz and Bystro\'nski, Mateusz and Holysz, Mikolaj and Binkowski, Jakub and Chodak, Grzegorz and Kajdanowicz, Tomasz},
  booktitle={Proceedings of the 63rd Annual Meeting of the Association for Computational Linguistics: Student Research Workshop},
  year={2025}
}

@inproceedings{kwon2023efficient,
  title={Efficient Memory Management for Large Language Model Serving with {PagedAttention}},
  author={Kwon, Woosuk and Li, Zhuohan and Siyuan, Zhang and Sheng, Ying and Zheng, Lianmin and Yu, Cody Hao and Gonzalez, Joseph and Zhang, Hao and Stoica, Ion},
  booktitle={Proceedings of the 29th ACM Symposium on Operating Systems Principles},
  pages={611--626},
  year={2023}
}

@inproceedings{yu2022orca,
  title={Orca: A Distributed Serving System for Transformer-Based Generative Models},
  author={Yu, Gyeong-In and Jeong, Joo Seong and Kim, Geon-Woo and Kim, Soojeong and Chun, Byung-Gon},
  booktitle={16th USENIX Symposium on Operating Systems Design and Implementation (OSDI 22)},
  pages={521--538},
  year={2022}
}

@article{zhong2024distserve,
  title={DistServe: Disaggregating Prefill and Decoding for Goodput-Optimized Large Language Model Serving},
  author={Zhong, Yinmin and Liu, Shengyu and Chen, Junda and Hu, Jianbo and Zhu, Yibo and Liu, Xuanzhe and Jin, Xin and Zhang, Hao},
  journal={arXiv preprint arXiv:2401.09670},
  year={2024}
}

@article{jiang2023prompteol,
  title={Scaling Sentence Embeddings with Large Language Models},
  author={Jiang, Ting and Huang, Shaohan and Luan, Zhongzhi and Wang, Deqing and Zhuang, Fuzhen},
  journal={arXiv preprint arXiv:2307.16645},
  year={2023}
}

@article{lei2024meta,
  title={Meta-Task Prompting Elicits Embeddings from Large Language Models},
  author={Lei, Yibin and Wu, Di and Zhou, Tianyi and Shen, Tao and Cao, Yu and Tao, Chongyang and Yates, Andrew},
  journal={arXiv preprint arXiv:2402.18458},
  year={2024}
}

@inproceedings{perez2018film,
  title={{FiLM}: Visual Reasoning with a General Conditioning Layer},
  author={Perez, Ethan and Strub, Florian and de Vries, Harm and Dumoulin, Vincent and Courville, Aaron},
  booktitle={Proceedings of the AAAI Conference on Artificial Intelligence},
  volume={32},
  number={1},
  year={2018}
}

@article{shazeer2017outrageously,
  title={Outrageously Large Neural Networks: The Sparsely-Gated Mixture-of-Experts Layer},
  author={Shazeer, Noam and Mirhoseini, Azalia and Maziarz, Krzysztof and Davis, Andy and Le, Quoc and Hinton, Geoffrey and Dean, Jeff},
  journal={arXiv preprint arXiv:1701.06538},
  year={2017}
}

@article{choi2025elis,
  title={{ELIS}: Efficient {LLM} Iterative Scheduling System with Response Length Predictor},
  author={Choi, Seungbeom and Goo, Jeonghoe and Jeon, Eunjoo and Yang, Mingyu and Jang, Minsung},
  journal={arXiv preprint arXiv:2505.09142},
  year={2025}
}

@inproceedings{wu2023fastserve,
  title={Fast Distributed Inference Serving for Large Language Models},
  author={Wu, Bingyang and Zhong, Yinmin and Zhang, Zili and Huang, Gang and Liu, Xuanzhe and Jin, Xin},
  booktitle={Proc.~{USENIX} Symp. on Networked Systems Design and Implementation ({NSDI})},
  year={2026}
}
